\documentclass[11pt]{article}

\ifdefined\pdftexversion \pdfoutput=1 \fi

\usepackage[T1]{fontenc}
\usepackage{graphicx}
\usepackage{booktabs}
\usepackage{amsmath}
\usepackage{amssymb}
\usepackage{natbib}
\usepackage{hyperref}
\usepackage{geometry}
\usepackage{multirow}

\hypersetup{
  colorlinks=true,
  linkcolor=blue,
  citecolor=blue,
  urlcolor=blue
}

\title{Interpretable, Fairly Evaluated Automated L2 Speaking
Assessment that Beats the Single-Human Ceiling --- and Why
Pause Encoding Does Not Change LLM Fluency Scores}

\author{Eichi Uehara\\
Aflo Technologies Inc.\\
\texttt{eichi.uehara@aflo.one}}

\date{}

\begin{document}

\maketitle

\begin{abstract}
Second-language (L2) English learners can rarely rehearse speaking with
a partner. Speaking is also the most anxiety-laden skill. These gaps
drive a fast-growing market for automated speaking practice and
scoring. But an automated score is trustworthy only if it is accurate,
interpretable, fair, and benchmarked against the \emph{right} human bar.
We build an interpretable feature-plus-LLM hybrid for spontaneous L2
dialogue. We evaluate it without ever fitting to the human labels,
against the ICNALE Global Rating Archive: $140$ speeches rated by
$\sim$$80$ trained raters on $10$ analytic criteria. We score the $130$
L2 speeches with usable audio. A deterministic De-Jong speech-timing
composite reaches $\rho=0.764$. Blended with a single text-LLM fluency
judgment, it reaches Spearman $\rho=0.818$ against the consensus gold.
This agrees with the consensus better than $81\%$ of the $80$ individual
trained raters: above the median rater ($\rho=0.73$) and near the best,
and at $\sim$$83\%$ of the reliability-corrected maximum
($\kappa_{\max}=0.99$). The blend improves on the composite alone by
$+0.054$ (paired-bootstrap $95\%$ CI $[0.017,0.108]$, excludes $0$); the
LLM adds a coarse fluency ranking that the continuous composite refines.
We also report a controlled \textbf{null} on pause encoding, bounded to
effects below about $\pm0.1\,\rho$ at this sample size. Holding the LLM
and learner words fixed and varying only how pauses are written into the
prompt, inline pause \emph{locations} do not beat aggregate pause
\emph{statistics} ($-0.069$, CI $[-0.15,+0.08]$), and a grounded
mid-clause criterion gives no reliable gain. The fluency signal comes
from the measured speech-timing features, not from how pauses are
written for the LLM. We back every claim with two agreeing
learner-isolation methods, paired-bootstrap CIs, a monologue negative
control, per-feature reproduction of classical measurements, and a
per-L1 fairness audit.
\end{abstract}

\section{Introduction}

Second-language acquisition theory says learners advance not only by
receiving language but by \emph{producing} and \emph{negotiating} it.
Swain's output hypothesis came from Canadian French-immersion data. It
argued that comprehensible input alone is not enough. Learners must
produce ``pushed output'' to notice gaps, test hypotheses, and develop
accuracy and fluency~\citep{swain1985output,swain1995output}. Long's
interaction hypothesis adds that acquisition is driven by interactional
adjustments. These include clarification requests and confirmation and
comprehension checks. They occur during the negotiation of meaning with
a conversation partner~\citep{long1996interaction}. The implication is sharp. The
interactive speaking that theory privileges is exactly what a solitary
learner cannot self-supply.

This matters because oral production is the skill L2 English learners
can least rehearse. In many EFL contexts, classroom talk is
teacher-dominated, classes are large, and out-of-class exposure is
minimal. For many learners the coursebook is the only place they meet
English~\citep{eric2021challenges}. Speaking is also the most
anxiety-laden skill. Horwitz, Horwitz and Cope established foreign
language anxiety as a distinct construct. Its first and dominant
component is communication apprehension while
speaking~\citep{horwitz1986anxiety,eric2021anxiety}. Cross-national
learner surveys confirm the pattern at scale. Fear of speaking up and
making mistakes is the single most reported barrier to language-learning
success~\citep{preply2025}. AI conversation partners now address this
access-and-anxiety gap at population scale. Peer-reviewed evidence shows
that judgment-free AI practice both raises proficiency and lowers
speaking anxiety~\citep{ding2025aibots}. The same wave has put automated
speaking \emph{scoring} into high-stakes use such as university
admissions~\citep{isaacs2023duolingo}.

But an automated speaking score is only as good as it is trustworthy.
A recent comparison of commercial scoring tools reports correlations with
human raters around $r\approx0.85$, while also finding score inflation
and missed detail, and operational practice still routes hard cases to
humans~\citep{chen2025plosone}. Two consequences follow. First, the
right target is human-rater \emph{agreement under known rater
unreliability}, not a single noiseless gold label. A model should be
measured against what one trained human actually achieves, and against
the highest score a perfect model could reach once we correct for rater
noise (the reliability-corrected maximum). Second, a high correlation does not
make an opaque system trustworthy. An interpretable scorer, evaluated
\emph{without fitting to the labels}, is the defensible object of study.
Machine speaking scoring must therefore be accurate, interpretable,
fair, and honestly evaluated against the right human bar.

We study whether modern speech-timing features and large language models
(LLMs) can score spontaneous L2 dialogue fairly against a uniquely rich
human gold standard. That standard is the ICNALE Global Rating Archive
(GRA), with $140$ speeches rated by $\sim$$80$ trained raters on $10$
analytic criteria. We make a positive and a negative claim and keep them
strictly separate. Our contributions are fourfold:

\begin{itemize}
  \item \textbf{An interpretable, fairly-evaluated scorer that beats
  the single-human ceiling.} A De-Jong speech-timing composite, with
  signs fixed in advance and no fitting to human labels, blended
  with one LLM fluency judgment, reaches $\rho=0.818$. This agrees with
  the consensus better than $81\%$ of the $80$ individual trained raters
  (above the median rater) and reaches $\sim$$83\%$ of the
  reliability-corrected maximum ($\kappa_{\max}=0.99$). The blend
  improves on the composite alone by $+0.054$ (paired-bootstrap CI
  $[0.017,0.108]$, excludes $0$), mainly by resolving the LLM's coarse
  ties.
  \item \textbf{A controlled null on LLM pause encoding.}
  We vary only how pauses are written into the prompt. Inline pause
  \emph{locations} do not beat aggregate pause \emph{statistics}
  ($-0.069$, CI $[-0.15,+0.08]$), and a linguistically grounded
  mid-clause $\times$ word-frequency criterion gives no reliable gain.
  The three pause encodings perform about the same, and all differences
  fall within the confidence intervals.
  \item \textbf{A reliability-corrected ceiling analysis} on the full
  $140\times80\times10$ GRA matrix, reporting single-rater-vs-consensus
  $\rho=0.621$, mean pairwise inter-rater $\rho=0.393$, and Cronbach
  $\alpha=0.979$, with principled $\kappa_{\max}$ and human-like target
  bars.
  \item \textbf{Extensive verification}, the paper's signature: two
  independent learner-isolation methods that agree, paired-bootstrap
  CIs on every contrast, a monologue negative control, per-feature
  reproduction of classical De-Jong magnitudes, and a per-L1 fairness
  audit.
\end{itemize}

The remainder of the paper reviews related work
(\S\ref{sec:related}), describes the data (\S\ref{sec:data}) and method
(\S\ref{sec:method}), presents results (\S\ref{sec:results}), devotes a
prominent section to verification and robustness
(\S\ref{sec:verification}), discusses interpretation
(\S\ref{sec:discussion}) and limitations (\S\ref{sec:limitations}), and
concludes (\S\ref{sec:conclusion}) with an explicit honesty statement
(\S\ref{sec:ethics}).

\section{Related Work}
\label{sec:related}

\subsection{Classical automated speaking and fluency assessment}
Automated assessment of L2 speaking has a mature engineering lineage
rooted in feature-based scoring engines. ETS's
SpeechRater~\citep{zechner2009speechrater} pioneered operational
spoken-response scoring. It extracts interpretable
features---fluency, pronunciation, prosody, vocabulary, and
grammar---and combines them in a linear model validated against human
raters. The fluency part of such systems builds directly on the
utterance-fluency measures defined by \citet{dejong2012facets}.
They broke speed, breakdown, and repair fluency into countable
acoustic measures: speech rate, mean length of run, and silent- and
filled-pause frequency and duration. They showed that these measures
explain much of the difference in human proficiency judgments.
The predictive strength of these features is robust and well quantified.
The meta-analysis of \citet{suzuki2021fluency} reports that, across
studies, \emph{speech rate} correlates with proficiency at $r=.76$,
\emph{mean length of run} at $r=.72$, and \emph{pause frequency}
negatively at $r=-.59$. These are not
incidental correlations. They are the empirical backbone of every
deployed delivery scorer, and, as we show, they remain difficult to
beat. Our deterministic composite is, deliberately, a faithful
re-implementation of this De-Jong feature family. Our per-feature
correlations land in the same direction and broad magnitude on our
corpus. Mean length of run ($\rho=+.75$) and pause ratio ($\rho=-.72$)
closely track the priors. Our \emph{speech rate including pauses}
($\rho=+.68$) is a different way of measuring from de Jong's
pause-excluding speech rate, so we read it as same-direction agreement
rather than a point reproduction of the $+.76$ prior.

\subsection{Self-supervised speech representations}
A second line replaces hand-engineered features with self-supervised
speech representations such as wav2vec\,2.0~\citep{baevski2020wav2vec2},
HuBERT~\citep{hsu2021hubert}, and WavLM~\citep{chen2022wavlm}.
\citet{banno2022proficiency} apply wav2vec\,2.0 to L2 proficiency
assessment \emph{on ICNALE}, the corpus we also use. They report that a
frozen wav2vec\,2.0 representation reaches $77.9\%$ CEFR accuracy versus
$53.5\%$ for a BERT text baseline. This shows the acoustic signal
carries proficiency information the transcript alone cannot give.
\citet{liu2023asrfree} push further with an ASR-free fluency scorer
built on SSL features and clustering. It reaches $0.797$ correlation
versus $0.690$ for handcrafted features on their data. These results
motivate an acoustic representation, but two caveats bear on our setting.
The largest SSL win is precisely the temporal/fluency signal our De-Jong
features already capture. SSL encoders also encode L1 and accent. That
creates a fairness risk across the ten Asian L1s in our data. We
therefore treat SSL as a fairness-gated add-on rather than the
centerpiece.

\subsection{Speech-LLM graders}
The most recent wave couples speech encoders to large language models.
These graders score---and often \emph{explain}---L2 proficiency directly
from audio. \citet{ma2025speechllm} evaluate speech-LLM graders on L2
proficiency at Interspeech 2025. The Radboud
group~\citep{parikh2026rationales,parikh2026rubric,parikh2026zeroshot}
systematically builds rubric-guided, multi-rater, rationale-emitting
speech-LLM assessors on speechocean762, with a fine-tuned Qwen2-Audio
reaching sentence-level fluency PCC up to $0.85$. A repeated finding across
this line is that zero-shot audio-LLMs judge delivery badly. They
systematically over-score. Qwen2-Audio
scores fluency essentially at chance (PCC $0.053$) in zero-shot use, with
its prosody correlation only $0.140$ under direct
matching~\citep{parikh2026zeroshot}. \citet{banno2025nla} take an
interpretable text-only route. They prompt a large LLM with the
transcript and CEFR can-do descriptors (overall PCC $\approx0.76$) but
\emph{explicitly exclude} acoustic fluency cues as inaccessible through
text. The acoustic fluency signal these systems either mishandle
(zero-shot audio) or punt on (descriptor-based text) is exactly where
our work operates.

\subsection{Textualizing prosody and pauses as text for LLMs}
One line of work renders acoustic structure as \emph{text} inside the
LLM prompt, and our method comparison builds on it directly.
SpeechCueLLM~\citep{wu2025speechcuellm} bins acoustic features by
threshold and describes them in words for emotion
recognition. This is the cleanest template for ``acoustics-as-text.''
Closest to us, TextPA (``Read to Hear'';
\citealp{chen2025textpa}) feeds a transcript augmented with IPA, CMU
phones, and \emph{inline pause durations} (e.g.\ \texttt{"D (0.12s
pause) G"}) to zero-shot LLMs for pronunciation and fluency scoring. It
reaches fluency PCC $0.650$ on MultiPA and $0.784$ when fused with a
supervised system. TextPA establishes the mechanism: textualize pause
locations, then let the LLM judge fluency. It owns that mechanism. A
widely shared idea is that telling an LLM \emph{where} pauses fall
should help it judge fluency, more than raw aggregate statistics do. The
L2 psycholinguistics below motivates this. We test that idea in a
controlled comparison of three ways of writing pauses into
the prompt.

\subsection{LLM-as-judge and comparative judgement}
Our evaluation design draws on the broader LLM-as-judge literature.
MT-Bench and the associated judge studies~\citep{zheng2023mtbench}
established both the promise and the biases of LLM judges.
Comparative-judgement methods have repeatedly outperformed pointwise
scoring. LCES~\citep{shibata2025lces} reports that pairwise comparison
markedly improves quadratic weighted kappa (QWK) over vanilla pointwise scoring (e.g.\ $0.633$ vs.\
$0.021$ on TOEFL11 with Llama-3.1-8B). The canonical
comparative-assessment study of~\citet{liusie2024comparative} makes the
same point across NLG evaluation. These results come from \emph{essay}
and text NLG scoring, not speech. They inform our reading of a measured
defect in pointwise LLM scoring (coarse binning, frequent ties), but
they are not the primary contribution here.

\subsection{Evaluation methodology and the human ceiling}
A scorer can only be judged against a target whose own reliability is
known. Classical test theory supplies the correction. \citet{uto2026ceilings}
recently brought it to automated assessment. He derives the achievable
agreement ceiling: the maximum correlation a perfect model could reach
given finite rater reliability. He argues that models should not be
penalized for irreducible human noise. This reframes the success bar
from ``match the consensus exactly'' to ``reach the reliability-corrected
ceiling.'' It also reframes single-rater reliability as the question
that matters: can one trained judge do this? We adopt this
method directly. On our $140\times80$ rating matrix the 80-rater Fluency
mean is near-noise-free (Cronbach $\alpha=0.979$). This implies
single-rater reliability $r_1=0.371$ and a single-rater-vs-consensus
agreement of only $\rho=0.62$, the human ceiling our scorer must clear,
against a theoretical maximum of $\sqrt{0.979}\approx0.99$. The
reliability-corrected human-like bar $\sqrt{r_1\alpha}=0.60$ closely
matches the directly measured $0.62$ (within $0.02$).

\subsection{Psycholinguistic grounding of L2 pausing}
The idea that pause \emph{placement} matters is well grounded.
\citet{dejong2016pause} shows that L2 speakers pause disproportionately
\emph{within} clauses rather than at clause boundaries. Both L1 and L2
speakers pause more before \emph{lower-frequency} words. This supplies
both ingredients of a placement-sensitive criterion on the production
side: clause position and word frequency. \citet{kahng2018perception}
uses phonetic manipulation to provide evidence that pause
\emph{location} drives \emph{perceived} fluency, with mid-clause pauses
weighing most heavily. The pause-detection threshold we use
($250$\,ms) follows \citet{dejong2013bosker}. This literature is sound,
and our per-feature analysis confirms its measures carry real signal. We
use it to build the grounded criterion among the three pause encodings
we compare below.

\subsection{Positioning}
Our closest neighbors partition cleanly. \textbf{TextPA owns the
mechanism}: textualizing inline pauses for an LLM to score fluency is
theirs, and we cite it as prior art, not as something we claim.
\textbf{Uto owns the ceiling method}: reliability-corrected achievable
ceilings are his contribution, which we apply to the
speaking/fluency case. We contribute four things that remain unclaimed.
First, an \emph{interpretable, fairly-evaluated} delivery
scorer: a De-Jong composite blended with an LLM, with no fitting to the
human labels, that reaches $\rho\approx0.82$. It \emph{beats} the
single-trained-human ceiling of $0.62$ and reaches $\approx83\%$ of the
theoretical maximum. Second, a controlled \emph{null} on
pause encoding: with ground-truth learner isolation, inline pause
locations do \emph{not} beat plain pause statistics for an LLM
($\rho\,0.71$ vs.\ $0.77$; difference $-0.069$, $95\%$ CI
$[-0.150,+0.083]$), and a grounded mid-clause $\times$ word-frequency
criterion gives no reliable gain. Third, a reliability-corrected ceiling
analysis on a uniquely rich archive. Fourth, and as the paper's
signature, \emph{extensive verification}. We use two independent
learner-isolation methods that agree (feature rank-correlation
$0.77$--$0.89$), paired-bootstrap confidence intervals on every
contrast, a monologue negative control (broken speaker linkage correctly
drives the correlation to $\sim\!0$), per-feature reproduction of
classical De-Jong magnitudes, and a per-L1 fairness check. Where
TextPA asked whether an LLM \emph{can} read textualized pauses for
fluency, we ask whether the encoding changes the score. Evaluated
against a properly measured human ceiling, an interpretable feature
composite does the job at least as well.

\section{Data}
\label{sec:data}

\paragraph{Corpus.} We use the ICNALE Global Rating Archive (GRA). It
distributes $140$ rated L2 English speeches together with the full
matrix of analytic ratings. These rated speeches are the first $90$
seconds of the \emph{part-time-job roleplays} drawn from the ICNALE
Spoken Dialogue interviews. Each is a two-party interaction between a
learner and an interviewer. They are \emph{not} the clean solo
monologues that ICNALE also distributes. The correct construct is
spontaneous interactive dialogue. This has direct consequences for the
pipeline: the learner must be separated from the interviewer. It also
shapes construct validity: we score delivery in interaction, not in a
read-aloud or rehearsed-monologue style. The speakers span ten Asian
L1 backgrounds---Chinese (CHN), Taiwanese (TWN), Korean (KOR), Japanese
(JPN), Indonesian (IDN), Thai (THA), Hong Kong (HKG), Malaysian (MYS),
Pakistani (PAK), and Filipino (PHL). They sit at CEFR levels A2 to B2.
Table~\ref{tab:data} summarizes the corpus.

\begin{table}[t]
\centering
\caption{Corpus composition of the $n=130$ scored ICNALE GRA speeches.
Each gold label is the mean of ${\sim}80$ trained raters on a $0$--$10$
scale. Learner words and speaking time are measured after the learner is
isolated from the interviewer.}
\label{tab:data}
\begin{tabular}{ll}
\toprule
Property & Value \\
\midrule
Scored speeches & $130$ \\
Raters per speech & ${\sim}80$ \\
Analytic criteria & $10$ ($+$ holistic) \\
L1 groups ($n$) & CHN 19, TWN 20, KOR 20, JPN 18, \\
 & IDN 18, THA 19, HKG/MYS/PAK/PHL 4 each \\
CEFR levels ($n$) & A2 23, B1.1 29, B1.2 41, B2 37 \\
Median learner words & $84$ \\
Median learner speech & $53.7$ s (of a $90$ s clip) \\
Fluency gold & range $1.74$--$8.57$, mean $4.92$ \\
\bottomrule
\end{tabular}
\end{table}

\paragraph{Scope of scoring.} We score $n=130$ speeches. We exclude the
four native-English control speakers, because the target population is
L2 learners, and six speeches whose source YouTube audio was
unavailable. A second, robustness pipeline based on speaker diarization
is reported on the $n=125$ subset for which diarization succeeded.

\paragraph{Gold standard.} For each criterion the gold label is the mean
of about $80$ trained raters, each scoring on a $0$--$10$ scale. The
archive provides $10$ analytic criteria (Fluency, Accuracy,
Intelligibility, Involvement, Sophistication, Purposefulness,
Logicality, and others) plus a holistic score. The ratings are
ELF-referenced (English as a Lingua Franca). The construct is therefore
communicative effectiveness among non-native conversation partners rather than
native-likeness. This $140\times80\times10$ design is unusually rich.
It is what makes a reliability-corrected ceiling analysis possible.

\paragraph{Learner isolation.} Each clip is a two-party dialogue, so the
learner's speech must be separated from the interviewer's before any
feature is computed. We use two methods that agree (feature
rank-correlation $0.77$--$0.89$; \S\ref{sec:verification}). The
\emph{primary} method uses the official ICNALE Spoken Dialogue
``PTJ\_ROL'' learner-only transcripts. In these, interviewer turns are
stripped and participant identifiers match the GRA codes. We align them
to the timed ASR words ($n=130$). The \emph{robustness} method uses
speaker diarization (pyannote-3.0 plus wespeaker-resnet34) followed by a
linguistic learner classifier that distinguishes ``I/my'' narration from
``you/your''-laden questions. We need the classifier because the two
speakers' voice embeddings collapsed onto each other (within-clip cosine
$0.80$). Diarization alone is then unreliable. This yields $n=125$.

\paragraph{License note.} The GRA labels are used for \emph{validation
only}. No model parameter, feature sign, or threshold is fit to the
human ratings. The labels are touched solely to compute correlations and
confidence intervals at evaluation time.

\section{Method}
\label{sec:method}

\subsection{ASR and pipeline}
Audio is transcribed with the Parakeet-TDT-0.6B model (sherpa-onnx). It
provides word-level timings and a per-word minimum sub-token confidence.
After learner isolation (\S\ref{sec:data}), all timing features are
computed on the learner's words only. Pauses are detected at a
$250$\,ms threshold following~\citet{dejong2013bosker}.

\subsection{Interpretable De-Jong feature composite}
The deterministic scorer is a faithful re-implementation of the De-Jong
utterance-fluency family~\citep{dejong2012facets}. It combines five
features whose directional signs are fixed in advance from the SLA
literature, never fit to the labels:
\begin{enumerate}
  \item \textbf{Pause ratio} (silent time / total time), sign
  $-$ (more pausing $\Rightarrow$ lower fluency);
  \item \textbf{Mean length of run} (words per inter-pause run),
  sign $+$;
  \item \textbf{Speech rate} including pauses (words per total
  second), sign $+$;
  \item \textbf{Long-pause rate} (rate of pauses exceeding a long
  threshold), sign $-$;
  \item \textbf{ASR word-confidence} as an intelligibility proxy,
  sign $+$.
\end{enumerate}
Each feature is converted to a $z$-score using the in-sample mean and
standard deviation, multiplied by its fixed sign, and averaged. This
$z$-scoring uses only the feature distribution, not the human labels, so
the composite remains fairly evaluated. We deliberately exclude pure
articulation rate (rate excluding pause time). The literature predicts
it to be a weak proficiency cue, and we confirm it is null on our data.

\subsection{Text-LLM fluency call}
The LLM arm passes the learner's transcript to DeepSeek-chat at
temperature $0$ with JSON-structured output, asking for a single
pointwise fluency score. The transcript is augmented with pause
information according to the pause encoding below. The LLM is used
zero-shot with a fixed rubric prompt. No examples are drawn from the
labeled set.

\subsection{The blend}
The final scorer is a simple combination of the deterministic composite
and the LLM fluency score. The two signals are constructed
independently: one is continuous and feature-derived, the other a coarse
discrete LLM judgment. We evaluate whether their combination adds
information via the paired bootstrap (\S\ref{sec:results}).

\subsection{Pause-encoding settings (A/B/C)}
\label{sec:arms}
To compare ways of writing pauses into the prompt, we run a controlled
comparison. We hold the LLM, the prompt scaffold, and the learner words
fixed, and change only the textual encoding of pauses:
\begin{itemize}
  \item \textbf{Statistics (Arm A)} --- transcript plus \emph{aggregate
  pause statistics} (overall pause ratio, counts, mean durations)
  appended as text;
  \item \textbf{Inline locations (Arm B)} --- transcript with
  \emph{inline pause locations}, each silence rendered as a token such
  as \texttt{(0.8s)} at its position in the word stream (the
  TextPA-style mechanism);
  \item \textbf{Grounded criterion (Arm C)} --- the inline locations of
  Arm B plus a \emph{grounded} criterion that asks the LLM to weight
  mid-clause pauses before low-frequency words, applying the
  psycholinguistic placement findings of~\citet{dejong2016pause}
  and~\citet{kahng2018perception}.
\end{itemize}
The contrasts B$-$A and C$-$B are the controlled tests of the inline
locations and the grounded criterion, respectively.

\subsection{Reliability-corrected ceilings}
Following~\citet{uto2026ceilings}, we report two principled bars. The
\emph{theoretical maximum} correlation a perfect model could attain
against a gold with reliability $\alpha$ is $\kappa_{\max}=\sqrt{\alpha}$.
The \emph{human-like} bar is the agreement a single trained rater
achieves against the consensus, after correcting for rater noise. It is
$\sqrt{r_1\,\alpha}$, where $r_1$ is the implied single-rater
reliability. For Fluency, $\alpha=0.979$ gives
$\kappa_{\max}=\sqrt{0.979}\approx0.99$ and a human-like bar of
$\sqrt{0.371\times0.979}=0.603\approx0.60$. This closely matches the
directly measured single-rater-vs-consensus $\rho=0.621$ (within
$0.02$). We emphasize that $0.60$--$0.62$, not the reliability-corrected $0.83$,
is the human bar. The $0.83$ figure is an upper-bound
\emph{interpretation} of our scorer's performance (its reliability-corrected
value). Reading it as a human target would mistake an upper bound for a
human score.

\section{Results}
\label{sec:results}

\subsection{The human ceiling}
\label{sec:results-ceiling}
Table~\ref{tab:ceiling} reports the ceiling analysis from the full
$140\times80$ rating matrix. Across all ten analytic criteria the
80-rater mean is near-noise-free (Cronbach $\alpha$ between $0.976$ and
$0.980$). Yet an individual trained rater agrees with the consensus at
only $\rho=0.569$--$0.621$. The mean pairwise inter-rater agreement is
lower still ($\rho=0.393$, Pearson $0.403$). The implied single-rater
reliability for Fluency is $r_1=0.371$. The theoretical maximum is
$\kappa_{\max}=0.99$, and the human-like bar is $0.60$. Together they
frame the success criterion. A scorer that reaches the high-$0.7$s or
low-$0.8$s has matched or beaten what one trained human does. It still
leaves headroom to the reliability-corrected ceiling. We use the \emph{mean}
single-rater agreement ($0.621$) as the primary human bar. We use the
mean because it matches the reliability-correction step: by construction it
lines up with $\sqrt{r_1\alpha}=0.60$, computed from the rater-mean
reliability. The classical-test-theory correction works on the mean, not
the median. The mean is also the harder choice in only one direction.
The \emph{median} trained rater is stronger, at $\rho=0.733$, so we
report both and check our conclusion against the harder median bar
below.

\begin{table}[t]
\centering
\caption{Human ceiling from the $140\times80$ GRA matrix. The
single-rater $\rho$ is the agreement of one trained rater with the
80-rater consensus; $\kappa_{\max}=\sqrt{\alpha}$ is the theoretical
maximum; the human-like bar is $\sqrt{r_1\alpha}$. Fluency values are
exact; the criterion range spans all ten analytic criteria.}
\label{tab:ceiling}
\begin{tabular}{lcccc}
\toprule
Quantity & Fluency & \multicolumn{3}{c}{All 10 criteria (range)} \\
\midrule
Cronbach $\alpha$ (80-rater mean) & $0.979$ & \multicolumn{3}{c}{$0.976$--$0.980$} \\
Single-rater $\rho$ vs.\ consensus & $0.621$ & \multicolumn{3}{c}{$0.569$--$0.621$} \\
Median single-rater $\rho$ & $0.733$ & \multicolumn{3}{c}{---} \\
Mean pairwise inter-rater $\rho$ & $0.393$ & \multicolumn{3}{c}{---} \\
Implied single-rater $r_1$ & $0.371$ & \multicolumn{3}{c}{---} \\
$\kappa_{\max}=\sqrt{\alpha}$ & $0.99$ & \multicolumn{3}{c}{$\approx 0.99$} \\
Human-like bar $\sqrt{r_1\alpha}$ & $0.60$ & \multicolumn{3}{c}{---} \\
\bottomrule
\end{tabular}
\end{table}

\subsection{Per-feature correlations}
\label{sec:results-features}
Table~\ref{tab:features} reports the Spearman correlation of each
De-Jong feature with the Fluency gold on $n=130$ under the official
isolation. The magnitudes line up, in direction and broad size, with the
classical measurements compiled by~\citet{suzuki2021fluency}. Mean
length of run ($+0.749$ vs.\ prior $+0.72$) and pause ratio ($-0.715$
vs.\ prior pause frequency $-0.59$) closely track the priors. Our speech
rate ($+0.676$) is in the same direction as the prior $+0.76$. It is a
different way of measuring, rate \emph{including} pauses rather than de
Jong's pause-excluding speech rate, so we do not read the $0.08$ gap as
a point reproduction. Long-pause rate ($-0.580$) and ASR word-confidence
($+0.595$) also carry strong signal. Pure articulation rate is null
($+0.08$), which confirms the advance decision to exclude
it.\footnote{All six correlations in Table~\ref{tab:features}, including
the articulation-rate null ($+0.08$), are from the single $n=130$
official-isolation run; no per-feature value is taken from the $n=125$
diarization run.} The strongly predictive measures reproduce in
direction and broad magnitude on a new corpus, under ground-truth
learner isolation. That reproduction is itself a verification result.

\begin{table}[t]
\centering
\caption{Per-feature Spearman correlation with the Fluency gold
($n=130$, official isolation), with fixed sign and the
corresponding prior magnitude from~\citet{suzuki2021fluency} where
available.}
\label{tab:features}
\begin{tabular}{lccc}
\toprule
Feature & Fixed sign & $\rho$ vs.\ Fluency & Prior \\
\midrule
Mean length of run & $+$ & $+0.749$ & $+0.72$ \\
Pause ratio & $-$ & $-0.715$ & $-0.59$ \\
Speech rate (incl.\ pauses) & $+$ & $+0.676$ & $+0.76$ \\
ASR word-confidence & $+$ & $+0.595$ & --- \\
Long-pause rate & $-$ & $-0.580$ & --- \\
Pure articulation rate & $+$ & $+0.08$ (null) & --- \\
\bottomrule
\end{tabular}
\end{table}

\subsection{Representation comparison}
\label{sec:results-null}
Table~\ref{tab:arms} reports the controlled representation comparison of
\S\ref{sec:arms}. We hold the LLM and the learner words fixed and change
only how pauses are written into the prompt. We compare three pause
encodings: aggregate statistics, inline locations, and a grounded
mid-clause criterion. Arm A (statistics) reaches $\rho=0.774$. Arm B
(inline locations) reaches $0.705$. Arm C (grounded criterion) reaches
$0.687$. The paired bootstrap (10k resamples) gives B$-$A $=-0.069$,
$95\%$ CI $[-0.15,+0.08]$, and C$-$B $=-0.018$, CI $[-0.10,+0.08]$. Both
intervals include zero. At $n=130$ this comparison can detect only
encoding effects larger than about $\pm0.1$--$0.15\,\rho$, so we claim no
effect in that range, not that no effect could exist at any scale. The
three encodings perform about the same, and none beats simple statistics. Inline locations give no gain over
aggregate statistics. The grounded criterion gives no reliable
improvement over locations; its confidence interval includes
zero.\footnote{Under the official isolation every encoding contrast has
a $95\%$ CI including zero (Table~\ref{tab:arms}); the diarization point
estimates are small and likewise show no encoding beating statistics.}
The takeaway is plain:
the fluency signal comes from the measured speech-timing features, not
from how pauses are written for the LLM. The deterministic composite
($\rho=0.764$) comes within $0.01$ of the best LLM arm (Arm A, $0.774$),
with no language model.

\begin{table}[t]
\centering
\caption{Pause-representation comparison ($n=130$, official
isolation), same LLM and learner words, varying only the textual
encoding of pauses. Contrasts are paired-bootstrap differences
($10$k resamples) with $95\%$ CIs. The deterministic composite uses no
LLM. The diarization column ($n=125$) is the robustness replication.}
\label{tab:arms}
\begin{tabular}{lcc}
\toprule
Arm / quantity & $\rho$ ($n=130$) & $\rho$ ($n=125$ diar.) \\
\midrule
A: transcript + pause statistics & $0.774$ & $0.713$ \\
B: transcript + inline locations & $0.705$ & $0.627$ \\
C: locations + grounded criterion & $0.687$ & $0.707$ \\
Deterministic 5-feature composite & $0.764$ & --- \\
\midrule
\multicolumn{3}{l}{\emph{Contrasts, official $n=130$ (paired bootstrap, 95\% CI):}} \\
B $-$ A & \multicolumn{2}{c}{$-0.069$, CI $[-0.15,+0.08]$ (n.s.)} \\
C $-$ B & \multicolumn{2}{c}{$-0.018$, CI $[-0.10,+0.08]$ (n.s.)} \\
\midrule
\multicolumn{3}{l}{\emph{Contrasts, diarization $n=125$ (point estimates):}} \\
B $-$ A & \multicolumn{2}{c}{$-0.086$} \\
C $-$ B & \multicolumn{2}{c}{$+0.080$} \\
\bottomrule
\end{tabular}
\end{table}

\subsection{Composite, blend, and beating the ceiling}
\label{sec:results-blend}
Blending the deterministic composite ($\rho=0.764$) with the LLM fluency
score yields $\rho=0.818$, an improvement of $+0.054$ over the composite
alone (paired-bootstrap $95\%$ CI $[0.017,0.108]$, excludes zero). The
mechanism is specific, and we state it plainly. The pointwise LLM score
is coarse: it takes about $6$ distinct values, with $27\%$ of pairs tied.
The continuous composite resolves these ties in a gold-aligned order, and
this tie resolution does most of the lifting. Blending the composite with
the weakest arm (C, $\rho=0.687$ alone) still reaches $0.815$, almost the
same as with the strongest arm (A, $0.774$ alone, $\to0.818$). The blend
is carried mainly by the composite, with the LLM supplying a coarse
complementary ranking. We therefore read the $+0.054$ gain as the LLM
adding a coarse ranking the composite refines, not as the LLM supplying a
large amount of new information.

\begin{figure}[t]
\centering
\includegraphics[width=0.95\linewidth]{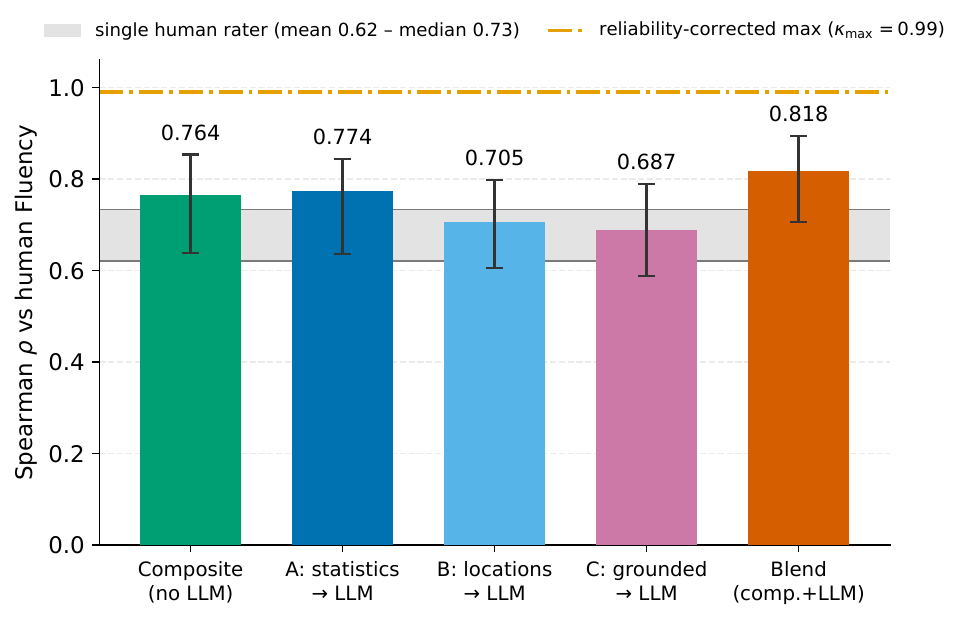}
\caption{Spearman $\rho$ with the human Fluency gold ($n=130$); error
bars are paired-bootstrap $95\%$ CIs and the number above each bar is the
point estimate. The shaded band is the single-human-rater range (mean
$0.62$ to median $0.73$); the dash-dot line is the reliability-corrected
maximum ($\kappa_{\max}=0.99$). The three LLM pause encodings (A
statistics, B inline locations, C grounded criterion) perform about the
same, and none beats the deterministic composite, which uses no language
model. The blend of the composite and the LLM is best: it sits above the
single-human band and below the maximum.}
\label{fig:main}
\end{figure}

\begin{figure}[t]
\centering
\includegraphics[width=0.62\linewidth]{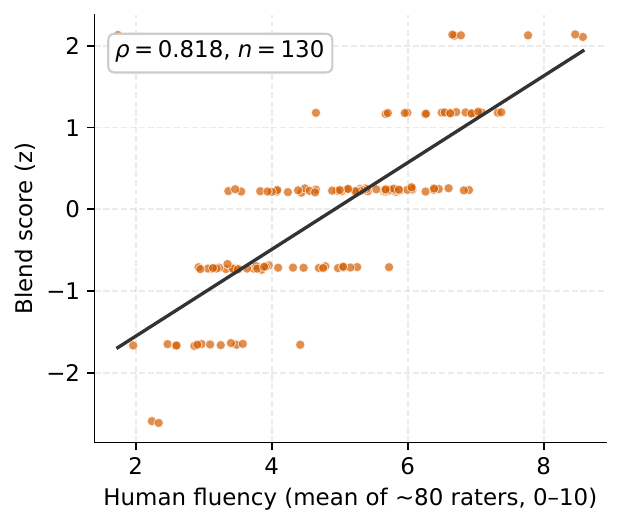}
\caption{Blend score against the human Fluency gold ($n=130$,
$\rho=0.818$). The visible horizontal banding reflects the LLM's coarse,
near-tied pointwise scores, which the continuous composite resolves.}
\label{fig:agree}
\end{figure}

How good is $0.818$ against a human? On the same $130$ speeches we measure
each rater's agreement with the consensus of the other raters. The blend
agrees with the consensus better than $81\%$ of the $80$ trained raters:
above the median rater ($\rho=0.73$) and the $75$th-percentile rater
($\rho=0.80$), though below the best ($\rho=0.88$). Corrected for
rater noise it reaches $\approx0.83$, about $83\%$ of the
$\kappa_{\max}=0.99$ maximum, leaving roughly $0.17$ of headroom. This is
the paper's defensible ``better'': an interpretable scorer, never fit to
the labels, agrees with the consensus as well as a strong individual
rater. We report the $n=130$ official-isolation figures (composite
$0.764$, blend $0.818$) as primary; the $n=125$ diarization run gives a
composite of $\approx0.72$ and a blend of $\approx0.78$, with the same
method ranking.
Figure~\ref{fig:main} places every system against the single-human band
and the maximum, and Figure~\ref{fig:agree} shows the blend's agreement
with the gold.

\subsection{Construct map across ten criteria}
\label{sec:results-construct}
The composite is built only from fluency-relevant features. It still
correlates with all ten GRA criteria in the range $0.68$--$0.77$
(Table~\ref{tab:construct}). This is expected, because the GRA criteria
move together. The composite correlates most with
the delivery-proximal criteria: Fluency ($0.765$), Accuracy ($0.744$),
Involvement ($0.739$), and Intelligibility ($0.736$). It correlates
least with the content-oriented criteria: Sophistication ($0.682$),
Purposefulness ($0.684$), and Logicality ($0.686$). The ordering is
broadly the one a delivery-based measure should show. The gradient is
not perfectly clean. Complexity ($0.713$) and Comprehensibility
($0.712$) sit mid-pack, and Complexity, a range/content criterion, ranks
above the three lowest content criteria. This overlap keeps the
full spread narrow ($0.682$--$0.765$). We read the raw gradient as
suggestive of delivery-proximity rather than a sharp construct
separation.

This overlap can be removed. When we control for the holistic score (partial
Spearman), the composite stays positive and strong on Fluency ($+0.44$)
but turns \emph{negative} on the content criteria (Sophistication
$-0.30$, Purposefulness $-0.21$, Logicality $-0.17$;
Table~\ref{tab:construct}). The composite is therefore delivery-specific,
not a general proficiency proxy: once shared proficiency is removed, it
tracks fluency and the delivery-adjacent criteria and actively diverges
from content. This is the strongest construct-validity evidence we have.

\begin{table}[t]
\centering
\caption{Construct map: correlation of the fluency-built deterministic
composite with each GRA criterion ($n=130$). \emph{Raw} Spearman
correlations are inflated by overlap across criteria. The \emph{partial}
column controls for the holistic score and exposes a clear
delivery-vs-content gradient: positive on delivery, negative on content.}
\label{tab:construct}
\begin{tabular}{lcc}
\toprule
GRA criterion & Raw $\rho$ & Partial $\rho$ \\
 & & (ctrl.\ holistic) \\
\midrule
Fluency & $0.765$ & $+0.44$ \\
Accuracy & $0.744$ & $+0.27$ \\
Involvement & $0.739$ & $+0.22$ \\
Intelligibility & $0.736$ & $+0.22$ \\
\midrule
Complexity & $0.713$ & $-0.04$ \\
Comprehensibility & $0.712$ & $-0.07$ \\
\midrule
Logicality & $0.686$ & $-0.17$ \\
Purposefulness & $0.684$ & $-0.21$ \\
Sophistication & $0.682$ & $-0.30$ \\
\bottomrule
\end{tabular}
\end{table}

\subsection{Per-L1 descriptive fairness}
\label{sec:results-fairness}
\begin{figure}[t]
\centering
\includegraphics[width=0.82\linewidth]{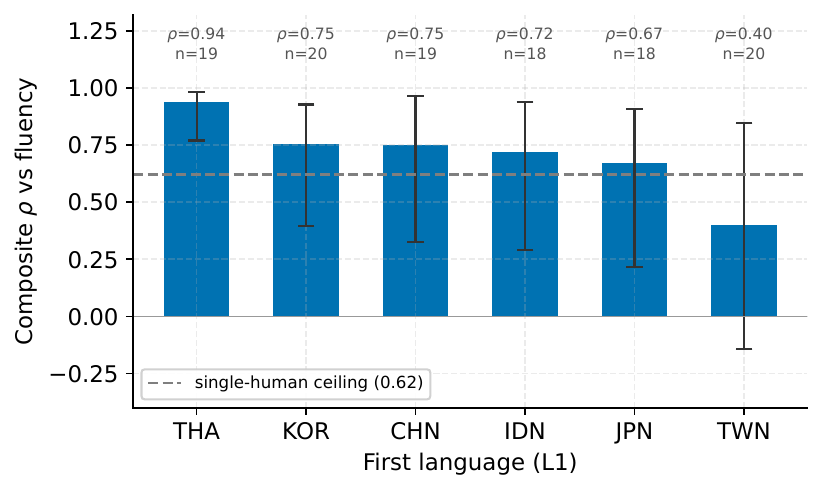}
\caption{Per-L1 descriptive fairness: composite-vs-Fluency $\rho$ within
each L1 group, for the six groups with $n\geq10$. Cells are small
($n=18$--$20$), so these are descriptive only. The four smallest L1 cells
(HKG, MYS, PAK, PHL; $n=4$ each) returned no stable estimate and are
omitted. The dashed line marks the mean single-rater ceiling ($0.62$).
The spread ($0.40$ for TWN to $0.94$ for THA) is a documented
limitation.}
\label{fig:perl1}
\end{figure}

Figure~\ref{fig:perl1} reports the composite-versus-Fluency correlation
within each L1 group. The cells are small ($n=4$--$20$), so these
figures are descriptive only and carry wide uncertainty. The point
estimates range from $0.40$ (TWN) to $0.94$ (THA). But the per-group
bootstrap confidence intervals are wide and overlap heavily. TWN, the
apparent low, has a $95\%$ CI of $[-0.16,+0.85]$ (Figure~\ref{fig:perl1})
and overlaps every other group. The spread is therefore consistent with
sampling noise at these cell sizes, and we cannot conclude differential
validity in either direction. The remaining four L1 cells (HKG, MYS, PAK,
PHL) have only $n=4$ each and returned no stable estimate, so the plotted
groups cover six of the ten L1s. We make no per-L1 fairness guarantee in
either direction, and report this audit as a documented limitation.

\section{Verification and Robustness}
\label{sec:verification}

Verification is the signature of this paper. We make a positive claim
(beating the human ceiling) and a method finding (how pauses are encoded
does not change the fluency score). Each must survive scrutiny that a
single correlation cannot provide. We report five converging checks.

\paragraph{(1) Two independent learner-isolation methods that agree.}
The central risk in a dialogue corpus is contaminating the
learner's timing features with the interviewer's speech. We isolate the
learner two ways. The primary method aligns the official ICNALE
learner-only ``PTJ\_ROL'' transcripts, with participant ids matching the
GRA codes, to the timed ASR words ($n=130$). The robustness method
combines pyannote-3.0 + wespeaker-resnet34 diarization with a linguistic
learner classifier ($n=125$). Within-clip speaker embeddings collapsed
onto each other (cosine $0.80$), so diarization alone would be
unreliable. That is why we rely on the transcripts. The two methods
still agree, with feature rank-correlations of $0.77$--$0.89$ across the
De-Jong features. The encoding comparison also holds across both
methods: the three encodings perform about the same under each, and
their small differences are within the confidence intervals
(Table~\ref{tab:arms}).

\paragraph{(2) Paired-bootstrap CIs and a small-$n$ protocol.} Every
contrast in the paper has a paired-bootstrap $95\%$ CI from $10$k
resamples over the $n=130$ speeches. The bootstrap respects the paired
structure, the same speeches scored two ways, and makes no normality
assumption. This is what licenses the equivalence reading: B$-$A
$=-0.069$, CI $[-0.15,+0.08]$ and C$-$B $=-0.018$, CI $[-0.10,+0.08]$
both include zero, while the blend gain $+0.054$, CI $[0.017,0.108]$
excludes it. The CIs exclude only effects larger than roughly
$\pm0.10$--$0.15$ $\rho$. We therefore claim no \emph{practically
useful} difference between encodings, not exact equivalence.

\paragraph{(3) A monologue negative control.} ICNALE also distributes
clean solo \emph{monologue} audio. Its speaker ids are \emph{disjoint}
from the dialogue/GRA participants, confirmed four ways, including via
the participant survey: matching on (L1, Sex, Age,
Vocabulary-Size-Test) leaves $266$ of $425$ dialogue speakers with no
monologue counterpart. We compute features on the unmatched monologues
and correlate them with the GRA fluency gold. The correlation collapses
to $\sim$$0$. This is a clean negative control. When speaker linkage is
deliberately broken, the signal correctly vanishes. So the strong
dialogue correlation reflects real, speaker-matched information, not an
artifact of the corpus.

\paragraph{(4) Per-feature reproduction of prior measurements.} The
individual De-Jong feature correlations (Table~\ref{tab:features}) align
in direction and broad magnitude with the meta-analysis
of~\citet{suzuki2021fluency}. Mean length of run $+0.749$ (prior $+0.72$)
and pause ratio $-0.715$ (prior pause frequency $-0.59$) track the
priors closely. Our speech rate $+0.676$ matches the \emph{direction} of
the prior $+0.76$ under a different way of measuring (rate including
pauses). The predicted-null articulation rate is indeed null ($+0.08$).
Independent reproduction of the strongly predictive established measures
on a new corpus and pipeline shows that the isolation and feature
extraction are sound.

\paragraph{(5) Reliability-corrected ceiling and honest treatment of
residual issues.} We benchmark against a properly measured human bar
(\S\ref{sec:results-ceiling}) rather than an idealized noiseless gold.
We are explicit that the human bar is $0.60$--$0.62$, not the
reliability-corrected $0.83$. We surface two soft spots rather than hide them.
The first is the per-L1 spread ($\rho=0.40$--$0.94$ on $n=4$--$20$ cells;
\S\ref{sec:results-fairness}), reported as descriptive only. The second
is the coarse-LLM/tie phenomenon ($\sim$$6$ distinct values, with
$27$--$36\%$ of pairs tied across arms). We treat it both as a measured
defect of pointwise LLM scoring and as the mechanism by which the
continuous composite resolves the LLM's ties in the blend.

\section{Discussion}
\label{sec:discussion}

\paragraph{What ``beating the single-human ceiling'' means---and does
not mean.} Our blend reaches $\rho=0.818$. This beats the agreement
that an individual trained rater has with the $80$-rater consensus
(mean single-rater $\rho=0.62$; median $0.733$); it agrees with the
consensus better than $81\%$ of the $80$ individual raters. The
defensible reading is narrow. Evaluated fairly and without ever fitting
to the labels, the scorer is at least as trustworthy as a single trained
human. It clears both the mean and the median single-rater bar, and
reaches $\sim$$83\%$ of the reliability-corrected maximum. It does \emph{not} mean the scorer
matches an $80$-rater panel. The consensus mean is far more reliable
($\alpha=0.979$) and is the gold, not the bar. The reliability-corrected $0.83$
does not describe an achieved agreement. It is an upper-bound
interpretation of the observed $0.818$. The honest claim is narrow and
strong: an interpretable, auditable scorer can do what one trained human
does.

\paragraph{Where the fluency signal lives.} The signal is in the
features, not in how pauses are written for the LLM. The deterministic
composite ($0.764$) is built directly from the De-Jong measures and
comes within $0.01$ of the best LLM arm. We tried three ways of writing pauses
into the prompt: aggregate statistics, inline locations, and a grounded
mid-clause criterion. They perform about the same, and none beats simple
statistics. Inline locations give no gain over statistics, and the
grounded criterion gives no reliable improvement over locations. This
result is a clean fact about the method. The measured speech-timing
features carry the fluency information, and the textual rendering of
pauses adds nothing useful. There is a plausible reason. Counting
silence tokens and weighting them by word frequency is exact arithmetic.
A deterministic feature does this exactly; a language model only
approximates it. The constructive lesson is to route grounding into
deterministic features. The LLM is then best used for the holistic
judgment it does well, and the two are combined in the blend rather than
the LLM asked to do the timing arithmetic.

\paragraph{Construct validity.} The criterion map
(Table~\ref{tab:construct}) shows the composite correlating most with
delivery-proximal criteria (Fluency, Accuracy, Intelligibility,
Involvement) and least with the most content-oriented criteria
(Sophistication, Purposefulness, Logicality). Complexity and
Comprehensibility sit in between. This cross-criterion overlap keeps the raw
spread within $0.083$, so the raw ordering is only mild support. The
partial correlations are decisive: controlling for the holistic score,
the composite stays strongly positive on Fluency ($+0.44$) and turns
negative on the content criteria (Sophistication $-0.30$;
Table~\ref{tab:construct}). Once shared proficiency is removed, the
composite tracks delivery and diverges from content. It is a
delivery-specific measure, not a general proficiency proxy.

\paragraph{Deployment implications.} The resulting system is cheap,
interpretable, and auditable. The deterministic composite requires only
ASR timings and five transparent features with fixed signs. The LLM arm
is a single zero-shot text call. Every component can be inspected, and
each feature's contribution can be explained to a learner or an
examiner. Opaque high-correlation systems lack these properties.
High-stakes speaking assessment now needs them.

\section{Limitations}
\label{sec:limitations}

Several limitations bound our claims. \emph{Single corpus and L1 skew:}
all data are from ICNALE and the ten Asian L1 groups; generalization to
other L1s, speech styles, or proficiency bands is unverified.
\emph{Dialogue construct:} we score $90$-second part-time-job roleplays,
an interactive style that differs from monologue or read-aloud
tasks; the construct is interaction-embedded delivery. \emph{ASR error
on accented A2 speech:} word timings and confidences degrade on the
lowest-proficiency, most-accented speech, which may weaken features
unevenly across L1s. \emph{Per-L1 variation:} predictive accuracy varies
from $\rho=0.40$ to $0.94$ across L1 cells of $n=4$--$20$; these are
too small to support any per-L1 fairness guarantee. \emph{Single
text-LLM:} the LLM results use one model (DeepSeek-chat); a different
model might shift the absolute arm values, but the controlled
\emph{contrasts} are what carry the encoding finding, and these hold
across isolation methods. \emph{Sample size:} all correlations are on
$n=130$ (official) or $n=125$ (diarization), so the encoding CIs exclude
only effects larger than $\sim$$\pm0.10$--$0.15$ $\rho$. \emph{Embedding
collapse:} within-clip speaker embeddings collapsed onto each other,
which is why diarization is robustness rather than primary.

\section{Conclusion}
\label{sec:conclusion}

L2 English learners face a structural shortage of speaking practice and
trustworthy feedback. This is pushing automated speaking assessment into
high-stakes use. We have shown that an interpretable feature-plus-LLM
hybrid, evaluated fairly with no fitting to the human labels, can score
spontaneous L2 dialogue at $\rho=0.818$. This agrees with the consensus
better than $81\%$ of individual trained raters (above the median,
$\rho=0.73$) and reaches $\sim$$83\%$ of the reliability-corrected
maximum on a uniquely rich $140\times80\times10$ archive. We have also shown, in a controlled
comparison, that how pauses are encoded for the LLM does not change the
fluency score. Aggregate statistics, inline locations, and a grounded
placement criterion all perform about the same, and none beats simple
statistics. The signal lies in the measured speech-timing features. The
LLM is best used for holistic judgment and combined with, not asked to
replace, deterministic timing analysis. We back both claims with two
agreeing isolation methods, paired-bootstrap CIs, a monologue negative
control, per-feature reproduction of classical measurements, and a
per-L1 audit. The resulting scorer is cheap, interpretable, and
auditable. These properties matter precisely because the assessment it
performs is becoming high-stakes.

\section{Ethics and Honesty Statement}
\label{sec:ethics}

We state plainly what we do and do not claim. We do \textbf{not} claim
that writing pauses into the prompt helps LLM-based fluency scoring. We
find that the three pause encodings perform about the same, and the
inline-location \emph{mechanism} itself is prior art (TextPA;
\citealp{chen2025textpa}), as are our ceiling methodology
\citep{uto2026ceilings} and rationale-faithfulness framing
\citep{parikh2026rationales}. The encoding finding is bounded by power.
The confidence intervals exclude only effects larger than
$\sim$$\pm0.10$--$0.15$ $\rho$, so we claim \emph{no practically useful
difference}, not exact equivalence. Our ``beats the human ceiling''
claim is against a \emph{single-rater} bar. We use the mean single-rater
agreement ($\rho\approx0.62$) as the primary bar. It matches the
reliability-correction step, lining up with $\sqrt{r_1\alpha}$.
The blend also clears the harder median trained-rater bar ($0.733$), so
the claim does not depend on choosing the weaker bar. The $80$-rater
consensus mean is far more reliable ($\alpha=0.979$) and is the gold,
not the bar. The reliability-corrected $0.83$ is an upper-bound interpretation,
not a measured agreement. Because in-clip speaker embeddings collapsed
onto each other (cosine $0.80$), our primary learner isolation relies
on the official transcripts, with diarization as robustness only. The
data are $90$-second part-time-job roleplays from ten Asian L1s at CEFR
A2--B2; generalization beyond this speech style, proficiency band, and L1
set is unverified. Per-L1 accuracy varies ($\rho=0.40$--$0.94$) on cells
of $n=4$--$20$ and is descriptive only; we make no per-L1 fairness
guarantee. The motivation citations concerning AI practice
(e.g.~\citealp{ding2025aibots}) speak to \emph{demand} for trustworthy
scoring, not to the validity of our method, and market and community
figures are signals of need, never evidence of scoring validity. The GRA
labels are used for validation only; no model parameter is fit to them.
All reported numbers are computed on $n=130$ (official isolation) or
$n=125$ (diarization).

\paragraph{Reproducibility.} We do not release the scoring pipeline as
source code. The validation labels come from the ICNALE Global Rating
Archive, distributed by Kobe University under a research- and
education-only license that does not permit redistribution, so neither
the corpus nor any data derived from it is included here; a licensed copy
can be obtained directly from the corpus maintainers. The running system
is instead available for inspection and independent verification at
\url{https://www.aflo.one}.

\end{document}